\documentclass[11pt,a4paper]{article}
\usepackage[hyperref]{emnlp2020}
\usepackage{times}
\usepackage{latexsym}

\usepackage{xcolor}

\usepackage{microtype}

\aclfinalcopy % Uncomment this line for the final submission
\title{Sound Natural: Content Rephrasing in Dialog Systems}

\author{Arash Einolghozati\thanks{equal contribution} \\
  Facebook \\
  \texttt{arashe@fb.com} \\
  \And
  Anchit Gupta\footnotemark[1] \\
  Facebook\\
  \texttt{anchit@fb.com} \\
  \And
    Keith Diedrick \\
  Facebook \\
  \texttt{kdiedrick@fb.com} \\
  \And
    Sonal Gupta \\
  Facebook \\
  \texttt{{sonalgupta@fb.com}} \\}

\date{}

\begin{document}
\maketitle
\begin{abstract}
We introduce a new task of rephrasing for a more natural virtual assistant. Currently, virtual assistants work in the paradigm of intent-slot tagging and the slot values are directly passed as-is to the execution engine. However, this setup fails in some scenarios such as messaging when the query given by the user needs to be changed before repeating it or sending it to another user. For example, for queries like `ask my wife if she can pick up the kids' or `remind me to take my pills', we need to rephrase the content to `can you pick up the kids' and `take your pills'. In this paper, we study the problem of rephrasing with messaging as a use case and release a dataset of 3000 pairs of original query and rephrased query. We show that BART, a pre-trained transformers-based masked language model with auto-regressive decoding, is a strong baseline for the task, and show improvements by adding a copy-pointer and copy loss to it. We analyze different trade-offs of BART-based and LSTM-based seq2seq models, and propose a distilled LSTM-based seq2seq as the best practical model.
\end{abstract}

\section{Introduction}

Virtual assistants have achieved very high accuracy in parsing queries for execution~\cite{gupta2018rnng}, such as reciting the weather or setting a reminder. However, in some scenarios, parsing alone is not enough to execute a request as expected. For example, when the user says “Tell Alice I’ll meet her in 10 minutes", executing the parsed message would send the tagged content “I'll meet her in 10 minutes" instead of a more appropriate message such as “I’ll meet you in 10 minutes”.
The other scenario where rephrasing is needed to better represent the user's request is when the user asks “Remind me to brush my teeth tonight". A more natural response would be “OK, I'll remind you to brush your teeth tonight".

To make a virtual assistant sound more natural, it needs to rephrase the user's query content before executing it. The task is different from paraphrasing, as we do not want to change the user's wording, i.e., the language formality or choice of words. Instead, we need to make minimal syntactic changes to make the utterance sound natural. As a use case, we work on the messaging domain, where we focus on rephrasing a message that needs to be inferred from the user query. This domain is so named as it covers requests about sending and receiving text and instant messages. Unlike the confirmation case, the message rephrasing is more complicated and can involve syntactic, pronoun or verb changes. Note that our goal is not to paraphrase the user's message but to rephrase it minimally, making it sound more natural when being sent to another user. As such, we need to maintain the semantics and style of the original content. 

Our contributions are as follows: (1) We introduce a new task and release a Message Content Rephrasing (MCR) dataset for this task consisting of 3k queries with tagged content and possible rephrases, (2) We explore various modeling approaches to achieve high accuracy on the MCR and modify existing pre-trained models to accommodate for the nature of this task, and (3) We show that distilling the pre-trained models into simple models can significantly close the performance gap.

\section{Data}
We first collected a task-oriented dataset of messaging utterances by asking our annotators to come up with natural scenarios in which a user wants to send a message to a second user. 

We observed that the collected queries contained two distinct types of messaging content: 1) where the content needs to be rephrased (REPHRASE) and 2) where the content should be used verbatim (EXACT). 
As such, the utterances were sent to another set of annotators to mark the message content in the utterances, disregard the ones that do not contain one, and mark whether the utterance belongs to the EXACT or REPHRASE class. Two annotators needed to agree on the labelling for this task, with a possible third for disagreement resolution. In the event of no resolution after three annotators, the query was reviewed individually.
Examples of each class is shown in Table~\ref{tab:examples} where the original content is tagged by brackets around it.

Next, we sent the utterances belonging to the REPHRASE class to a different set of annotators and asked them to rephrase the message content in a way that it would be natural to send to the second user without any additional context. 

During annotation, our goal was to minimally rephrase a sentence, e.g., keeping the words and attributes (e.g., formality) of the original content as much as possible. In order to ensure high quality, we asked three annotators to independently rephrase the utterances. In around $30\%$ of the cases, there was not a majority (i.e., two or more annotators agree) and we asked a fourth annotator to resolve. Most of the disagreements were due to changing words that did not need to be changed for minimal rephrasing but there were cases where the minimal rephrase was not obvious. We will discuss this further when introducing our metrics.

 Overall, we have around 3k examples (almost half for each class) which we split by 70/20/10 for train/test/validation, respectively.
We can see from the training data that rephrasing mostly involves making a question and/or changing the subject pronoun. There are other linguistically complex scenarios such as deciding when to use politeness strategies (e.g. “Could you pick up milk" as opposed to “Can you pick up milk?") among these queries as well. We decided that these complex edge cases were best addressed in future work. As such, we cluster the rephrasing into three main categories. In Table~\ref{tab:examples}, the first example only needs a pronoun change, the second needs the form to be changed to a question, and the third needs both. We have also put the statistics for the changes in table~\ref{tab:change_stat}.
\footnote{The dataset can be downloaded from \href{dl.fbaipublicfiles.com/rephrasing/rephrasing_dataset.tar.gz}{dl.fbaipublicfiles.com/rephrasing/rephrasing\_dataset.tar.gz}}
\begin{table*}
%\vspace{-0.5cm}
\centering
\begin{tabular}{lcc}
\hline
\textbf{Query} & \textbf{Query} & \textbf{Rephrase}\\
\hline
REPHRASE & Let Kira know [ I can pick her up ] &	 I can pick you up\\
REPHRASE & Message Donna and ask [ when dinner is ] &	 when is dinner\\
REPHRASE & message Brad and ask [ if he has my keys ] &	 do you have my keys\\
EXACT & Tell Jo [I will be on time] & I will be on time\\
EXACT & ask my boss [will I have to work on Friday] & Will I have to work on Friday\\
 \\\hline
\end{tabular}
\begin{tabular}{lc}
\hline
\end{tabular}
\caption{Examples of queries and the rephrased utterance}
\label{tab:examples}
\end{table*}

As mentioned earlier, there is a huge overlap between the source and target sequences. As such, rephrasing can be viewed as a post-editing task more than a generation task. In Table~\ref{tab:stats}, we have showed some basic statistics about the training data for the REPHRASE class in MCR. 

\begin{table}
\centering
\begin{tabular}{lcccc}
\hline
\textbf{ Source} & \textbf{Target} & \textbf{Keep} & \textbf{Add} & \textbf{Delete}\\
7.9 & 9.3 & 5.9 & 3.4 & 2.0\\
\hline
\end{tabular}
\caption{Average Length and overlap between source and target}
\label{tab:stats}
\end{table}

\begin{table}
\centering
\begin{tabular}{lc}
\hline
EXACT (no changes)& $57\%$ \\
REPHRASE (pronoun) & $8\%$ \\
REPHRASE (question)  & $13\%$ \\
REPHRASE (pronoun+question) & $22\%$ \\
\hline
\end{tabular}
\caption{Frequency of the needed changes}
\label{tab:change_stat}
\end{table}

\section{Evaluation}
Our goal is to maximize the rephrasing accuracy while also maintaining a very high accuracy on the EXACT class.
Our first metric is the Exact Match (EM) accuracy in which the predicted rephrase should be the same as the original content for the EXACT class and equal to the top rephrased candidate for the REPHRASE class. The downside of this metric is that for utterances such as ‘ask her to pick up her phone’, we would penalize rephrases such as ‘can you pick up your phone’ if the gold label was ‘pick up your phone’. In order to smooth this metric, we also use $\textrm{EM}_\textrm{any}$ in which the rephrased content is correct if it matches any of the provided annotations.

Since the required changes to rephrase the content are usually small, the BLEU score may not be useful. On the other hand, not all the wrong rephrases are equal, e.g., when the model hallucinates. Metrics such as BLEU can penalize these phenomena more than the EM metrics.
We also use SARI~\cite{SARI}, which is commonly used for text-editing tasks. It measures the average F1 score of three editing actions for ngrams: Keep, Add, and Delete. 

\section{Modeling Approaches}

 We assume that the gold tagging for the content inside the query is provided. 
Our base model is an LSTM seq2seq model with two-layers for both encoder and decoder using Glove~\cite{glove} initialized word embeddings (20k vocab size) concatenated with ELMo~\cite{elmo} embeddings to represent the tokens. 
We also use the pointer-generator mechanism~\cite{pointer-generator}, which can choose between copying from the source or generating new tokens using a pointer-attention mechanism. As we can see in Table~\ref{tab:generation}, the copy mechanism is crucial in our task, as most of the tokens are copied from the source.

% Copy pointer math
The copy pointer works as follows: We calculate two token output probabilities; one over the full vocab $P^t_{vocab}$ using the standard softmax and another $P^t_{copy}$ over the source tokens. To obtain $P^t_{copy}$ we use a learned attention between the decoder hidden $h^t_{d}$ and the encoder outputs $H_{e}^T$. To generate the output, we weigh between copying and generation using a parameter $\alpha_{mix}$ which is also computed as a function of the hidden states, i.e., $ P^t_{output}=(1-\alpha_{mix})P^t_{vocab}+\alpha_{mix}P^t_{copy}$.
More precisely:
$$q^t, K, V = h_{d}^t W^T_q, H_{e}^TW^T_k, H_{e}^TW^T_v$$
$$P_{copy}^t  = softmax(q^TK) $$
$$\alpha_{mix}^t = sigmoid(W_{mix}.concat(q^TK,V)),$$
where $q$, $K$, $V$ are the query, key and value, respectively, needed to calculate the attention and all the $W_{*}$ matrices are learned parameters.

\begin{table*}
%\vspace{-0.5cm}
\centering
\small
\begin{tabular}{lcccccc}
\hline
\textbf{Model} & \textbf{EM} & \textbf{$\textrm{EM}_{\textrm{any}}$}&  \textbf{BLEU}  &
\textbf{$\textrm{EM}_{\textrm{exact}}$}& \textbf{$\textrm{EM}_{\textrm{rephrase}}$}& \textbf{SARI}
\\
\hline
Exact Copy & 55.0 &  55.0 & 80.6 &100  &0 & 26.3  \\
\hline 
LSTM seq2seq & 84.1  & 85.8&  91.0 & 96.6 & 68.9  & 83.1\\
LSTM seq2seq w/o ELMo & 81.3 & 82.4 & 89.4 & 93.8 & 66.1 &81.3 \\
LSTM seq2seq w/o Copy & 54.7 & 55.8 & 78.9 &62.3 &39.2 & 69.7\\
\hline
BART &  88.2 & 90.5 & \bf{96.0} & 95.5 &79.2 & \bf{86.4} \\
BART w copy & \bf{89.3} & \bf{92.1}& \bf{96.1} & \bf{96.9} & \bf{80.0} & \bf{86.5}\\
\hline
LSTM + seq-level KD & 84.1 & 86.1& 90.5 &96.6& 68.9 & 82.9\\
LSTM + seq-level KD + FT & 85.4 &   87.2  & 94.0 & 95.5 &73.0 & 83.7\\
\hline
LaserTagger & 87.4 & 88.7 & 94.6 & \bf{97.2} & 75.8 & 84.0
\end{tabular}
\caption{Rephrasing Model Performance}
\label{tab:generation}
\end{table*}

We have shown the LSTM results alongside ablation on the ELMo and Copying mechanism in Table~\ref{tab:generation}. We can see that copying is crucial, especially for the EXACT class.
We show the results for copying the content part of the source in the first row.

We also experiment with using BART~\cite{bart} for this task. BART is a powerful pre-trained seq2seq model trained on a de-noising objective over massive amount of web data. The training details are listed in the Appendix.
During our initial experiments with BART, we realized it can replace proper nouns when rephrasing. Even though BART is a de-noising autoencoder and it has a high proclivity to copy the source through its encoder-decoder attention heads, it is still done over the whole vocabulary space (50k bpe tokens) and not the dozen of source tokens. To address this PEGASUS~\cite{pegasus} is pre-trained by generating a selected masked sentence from the input, where some of the selected sentences are not masked. We instead opt to add an explicit copying to BART in the fine-tuning stage.

Since the pre-trained model has no explicit copy mechanism, adding it naively during the fine-tuning phase as above is not effective. In this case, the decoder prefers to use the well-trained generator instead of a randomly initialized attention head for copying.
We use two strategies to mitigate this: (1) We initialize the copying attention head with the average of the last layer's pre-trained decoder attention head, and (2) We also add an explicit loss that forces the decoder to use the copying mechanism when it can.
For all the target tokens that can be found in the source, we add a hinge loss: $\lambda max( T - P , 0)$ to the cross-entropy loss which forces the copying probability $P$ for those token to be above a threshold $T$. Hyper-parameters
$\lambda$ and $T$ are optimized over the validation set, $0.25$ and $0.9$, respectively.

We show results using the BART large model in Table~\ref{tab:generation}. Vanilla BART yields strong results compared with the LSTM seq2seq model for the rephrasing class but slightly lags for $\textrm{EM}_{\textrm{exact}}$, which requires pure copying. 
On the other hand, by adding the explicit copying to BART, it significantly improves the accuracy for both classes. Moreover, the gap between $\textrm{EM}$ and $\textrm{EM}_{\textrm{any}}$, the biggest for BART, shows the proportion of errors due to subtle differences within the resolved annotation, as opposed to errors caused by serious problems such as hallucination. 

\begin{table*}
\centering
\small	
\begin{tabular}{l|ccc|c}
\hline
\textbf{Model} & \textbf{Semantic} & \textbf{Grammatical}  & \textbf{Copy Related} &    \textbf{Correct}  \\
\hline
BART & 4\% & 13\% & \bf{24\%}  & 59\%   \\
BART w Copy & 14\% & 10\% & 8\% & \bf{68\%}  \\
Distilled LSTM  &  8\%  & \bf{45\%} & 8\% &  39\%  \\
LaserTagger & \bf{25\%} & 38\%  &  4\% & 33\%\\
\hline
\end{tabular}
\caption{Prevalence of each category of the models' mispredictions 
 }
\label{tab:ea}
%\vspace{-.5cm}
\end{table*}

\subsection{Distilling BART}

Deploying models such as BART can be prohibitive for real-time applications. It has 514M parameters and around 10X average CPU inference latency compared with the LSTM model that has only 9.6M parameters. Unlike the pointer-generator LSTM model, BART with copying still exhibits an over-generation problem while the LSTM model makes many grammatical errors. As such, we look into Knowledge Distillation (KD)~\citep{kd-hinton} to transfer the language modeling capability of BART while keeping its copying behavior.
Transferring the language model of massive pre-trained models into smaller models has been of high interest recently~\citep{distilbert, wellread_kd, patientKD}. Knowledge transfer to simple models has also been discussed in lesser extent~\citep{KD_simple, KD_simple_unlabeled}. 
We use the sequence-level distillation introduced in~\citep{sequence-kd-kim} and train the LSTM model using the BART output. We found that fine-tuning on the gold labels after the KD step is also beneficial to the performance.

\subsection{Edit vs Generate}

In a pure generation framework, e.g., BART without the copying loss, all the tokens are generated from scratch. On the other side of the spectrum, models such as LaserTagger~\cite{lasertagger} keep the original utterance and try to edit by adding or removing as needed. Adding the copying mechanism to our models can be considered a middle ground between editing and generation.
We use the framework introduced in~\cite{lasertagger} to edit the queries. It tags each word as Keep or Delete plus the optional phrase that needs to be added before it. We procure the list of phrases that yield high coverage over the training data in MCR. By using the top 100 phrases, we get coverage over $95\%$ of the training data. Note that the verb conjugations needed in our problem can cause a lack of generalization when using such limited vocabulary.

We train a tagging model using the RoBERTa encoder~\cite{roberta} with one layer of MLP and CRF on top of it.
We have listed the editing model performance on the last line of the Table~\ref{tab:generation}.
We can see that the editing yields better EM than the LSTM model but worse than BART. It is unsurprisingly the best model when no rephrasing is needed. On the other hand, the type of rephrasing errors it makes may be worse than the generative models as evidenced by the lower SARI score. For example, we find grammatical errors such as “did you I leave my sunglasses there”. This is possibly caused by the added words being treated as categorical classes and not as words in a LM.

\subsection{Error Analysis}
We cluster the errors into three categories with an additional `Correct' class which means that the prediction is correct but does not match any of the gold annotation exactly. A prominent example of the latter is the addition of politeness prefixes such as `Could you' to the beginning of a request which we discussed earlier.
 
 The Grammatical error class represents cases where the semantics can be understood but there are some grammatical errors such as a mismatch between the noun and verb forms (e.g. verb tense not matching noun person or number). 
 In the Semantic error class, the meaning is seriously affected. Semantic errors cover two distinct sub-categories which both change the meaning of the message: hallucinating new content and omission of parts of the content.
 The Copy-related error category happens mostly for proper nouns that are not carried over as exact copies into the output. Since this is observed mostly in the vanilla BART, we decided to separate this category from the rest of the errors.
 
 Note that if there are multiple classes of errors in the output, we pick the most prominent type of error for that utterance. 
 In Table~\ref{tab:ea}, we have shown the prevalence of each Category. 
We can see that in BART models, the majority of the ostensible errors are actually correct but the BART model without the explicit copying has the biggest copy-related errors among all models. Moreover, while Grammatical errors is the biggest category in both the distilled LSTM and the LaserTagger, the latter makes many more semantic errors which echoes our qualitative observation.

% \begin{table}
% \centering
% \begin{tabular}{lccc}
% \hline
% \textbf{Model} & \textbf{EM} & \textbf{Latency (ms)}  \\
% \hline
% BART & 88 & 550 \\
% LSTM seq2seq & 84.1  & X \\
% LaserTagger (RoBERTa) & X & X \\
% \hline

% \end{tabular}
% \caption{Latency Analysis}
% \label{tab:latency}
% \end{table}

\section{Related Work}
\subsection{Pre-trained Models for Generation}
Pre-training transformers on massive amounts of unlabeled data has resulted in recent advances in language understanding and generation tasks~\cite{bert, gpt2}.  Pre-trained encoder-decoder models have unified the benefits for both discriminative and generative tasks through pre-training as de-noising auto-encoders~\cite{MASS, bart, t5}.
\cite{fewshotNLG} fine tune such a big pre-trained model and add a copy pointer for a few shot structured tabular data summarization task.
% More recently,~\cite{pegasus} has extended the de-noising to over multiple sentences.
%
\subsection{Paraphrasing} 
Paraphrase generation using seq2seq models~\cite{seq2seq} has been recently discussed in the literature. \citet{residual_paraphrase} used residual LSTM seq2seq networks to perform paraphrasing. Unlike paraphrasing, in MCR, preserving the semantics of a message is necessary but not enough. Instead, we make minimal changes to make the sentence sound natural.
 \subsection{Sentence Editing and Simplification}
Automatic post-editing is applied to paraphrases and machine translation~\cite{quickedit}. Similar to this is Grammatical Error Correction which seeks to correct errors such as grammar and punctuation~\cite{gec-conll, gec-2019}. Sentence revision~\cite{diamonds} extends this to cases for which major rewriting may be needed.
Sentence simplification~\cite{neural-simplification} aims at using techniques such as shortening the sentences to make a text more readable. On the other hand, style transfer is the task of making an utterance conform to a specific style such as formality~\cite{style-transfer-2018, sennrich-2016-politeness}. From this perspective, the rephrasing task can be viewed as changing the style from the third-person to the second-person language and/or forming a question.

\section{Conclusion}
In this paper, we introduce a new task of message rephrasing in task-oriented dialog. We release a dataset, MCR, for this task and propose a new model (BART with copy). We show that adding an explicit loss to a pre-trained generative model during fine-tuning can improve the copying performance without hurting its generation power. We also show that by distilling the pre-trained model into a much smaller LSTM seq2seq model with copy pointer, we can significantly improve the LSTM seq2seq model's language model capability while still keeping its accurate copying.

\bibliographystyle{acl_natbib}
\bibliography{emnlp2020}

\clearpage

\appendix

\section{Appendix}
Here, we describe the details regarding the training.
In Table~\ref{tab:training_params}, we have shown the training details for all of our models. We use ADAM~\cite{adam} with Learning Rate (LR), Weight Decay (WD), and Batch Size (BSz) values that are listed for each model. We have also shown the number of epochs and the average training time for the full CS data using $4$ V100 Nvidia GPUs.

In all of our BART experiments, we have used BART large from PyText\footnote{\url{https://pytext.readthedocs.io/en/master/xlm\_r.html}}~\cite{pytext}. 
When adding the copying loss to BART, we fine-tuned the hyper-parameters $\lambda$ and $T$ over [0.1, 1] and [0.5,1], respectively, with increments of $0.05$.
We also use the RoBERTa large from PyText for the LaserTagger experiment.

In our LSTM models, the encoder and decoder are 2-layer LSTMs with hidden dimension of $128$ and $256$, respectively. We also use dropout of $0.3$ for all connections. The ELMo and GloVe embeddings have dimesions of $512$ and $200$, respectively, and we use the top 8k words in GloVe as our vocabulary. 

% In Table~\ref{tab:generation_eval}, we have listed the validation results for the models described in the paper.

\begin{table}
\centering
\begin{tabular}{lccccc}

\hline
Model & BSz & LR & WD  & Epoch & Avg Time  \\
BART &  10 &. 0.00002 & e-5 & 10 & 1hr \\
LSTM seq2seq & 16 &  $0.001$ &  e-6 & 40 & 45min   \\
RoBERTa &  16 & 0.000005 & e-4 & 10 & 4hr \\
\hline
\end{tabular}
\caption{Training Parameters}
\label{tab:training_params}
\end{table}

% \begin{table*}
% %\vspace{-0.5cm}
% \centering
% \begin{tabular}{lcccccc}
% \hline
% \textbf{Model} & \textbf{EM} & \textbf{$\textrm{EM}_{\textrm{any}}$}&  \textbf{BLEU}  &
% \textbf{$\textrm{EM}_{\textrm{exact}}$}& \textbf{$\textrm{EM}_{\textrm{rephrase}}$}& \textbf{SARI}
% \\
% \hline
% Exact Copy & 57.0 &  57.0 & 82.3 & 100  & 0 & 25.5  \\
% \hline 
% LSTM seq2seq & 84.3  & 86.2&  91.4 & 96.9 & 68.7  & 83.3\\
% LSTM seq2seq w/o ELMo & 81.6 & 82.7 & 90.2 & 94.0 & 66.0 & 81.8 \\
% LSTM seq2seq w/o Copy & 53.7 & 54.6 & 77.8 & 61.0 & 38.7 & 69.7\\
% \hline
% BART     &          87.5 & 90.0 &   95.3 & 96.1 &  77.2 & 85.0 \\
% BART w copy & \bf{89.1} & \bf{91.0}& \bf{95.6} & \bf{97.4} & \bf{77.9} & \bf{85.8}\\
% \hline
% LSTM + seq-level KD & 84.0 & 85.9& 90.4 &96.5& 69.0 & 83.0\\
% LSTM + seq-level KD + FT & 85.5 &   87.4  & 94.3 & 95.8 &73.1 & 83.8\\
% \hline
% LaserTagger & 87.6 & 88.8 & 94.7 & 97.1 & 75.6 & 84.2
% \end{tabular}
% \caption{Rephrasing Model Performance for the validation set}
% \label{tab:generation_eval}
% \end{table*}

\end{document}